\documentclass[final]{cvpr}

\usepackage{epsfig}
\usepackage{color}
\usepackage{times,graphicx,amsmath,amssymb,booktabs,tabulary,multirow,overpic,xcolor}
\usepackage{subcaption}
\definecolor{citecolor}{RGB}{34,139,34}
\usepackage[british,english,american]{babel}

\graphicspath{ {figures/} }

\newcommand{\bd}[1]{\textbf{#1}}
\newcommand{\app}{\raise.17ex\hbox{$\scriptstyle\sim$}}

\newcolumntype{x}[1]{>{\centering\arraybackslash}p{#1pt}}
\newlength\savewidth

\makeatletter\renewcommand\paragraph{\@startsection{paragraph}{4}{\z@}
  {.5em \@plus1ex \@minus.2ex}{-.5em}{\normalfont\normalsize\bfseries}}\makeatother

\usepackage[pagebackref=true,breaklinks=true,colorlinks,bookmarks=false]{hyperref}

\def\confYear{CVPR 2021}
\newcommand{\vct}[1]{\boldsymbol{#1}} % vector
\newcommand{\mat}[1]{\boldsymbol{#1}} % matrix

\newcommand{\methodname}{{MTLFace}\xspace}

\begin{document}

%%%%%%%%% TITLE
\title{When Age-Invariant Face Recognition Meets Face Age Synthesis: \\A Multi-Task Learning Framework}

\author{Zhizhong Huang$^{1}$\qquad Junping Zhang$^{1}$\qquad Hongming Shan$^{2,3,*}$ \\
$^{1}$ Shanghai Key Lab of Intelligent Information Processing, School of Computer Science,\\
Fudan University, Shanghai 200433, China\\
$^{2}$ Institute of Science and Technology for Brain-inspired Intelligence and MOE Frontiers Center \\for Brain Science,  Fudan University, Shanghai 200433, China\\
$^{3}$ Shanghai Center for Brain Science and Brain-inspired Technology, Shanghai 201210, China\\
{\tt\small \{zzhuang19, jpzhang, hmshan\}@fudan.edu.cn}
}

\maketitle

%%%%%%%%% ABSTRACT

%!tex root=cvpr.tex

\begin{abstract}
   To minimize the effects of age variation in face recognition, previous work either extracts identity-related discriminative features by minimizing the correlation between identity- and age-related features, called age-invariant face recognition (AIFR), or removes age variation by transforming the faces of different age groups into the same age group, called face age synthesis (FAS); however, the former lacks visual results for model interpretation while the latter suffers from artifacts compromising downstream recognition. Therefore, this paper proposes a unified, multi-task framework to jointly handle these two tasks, termed \methodname, which can learn age-invariant identity-related representation while achieving pleasing face synthesis.
   Specifically, we first decompose the mixed face feature into two uncorrelated components---identity- and age-related feature---through an attention mechanism, and then decorrelate these two components using multi-task training and continuous domain adaption. In contrast to the conventional one-hot encoding that achieves group-level FAS, we propose a novel identity conditional module to achieve identity-level FAS, with a weight-sharing strategy to improve the age smoothness of synthesized faces. In addition, we collect and release a large cross-age face dataset with age and gender annotations to advance the development of the AIFR and FAS.
   Extensive experiments on five benchmark cross-age datasets demonstrate the superior performance of our proposed \methodname over existing state-of-the-art methods for AIFR and FAS. We further validate \methodname on two popular general face recognition datasets, showing competitive performance for face recognition in the wild. The source code and dataset are available at~\url{https://github.com/Hzzone/MTLFace}.
\end{abstract}
%!TEX root=cvpr.tex
\section{Introduction}
\begin{figure}[t]
    \centering
    \includegraphics[width=.75\linewidth]{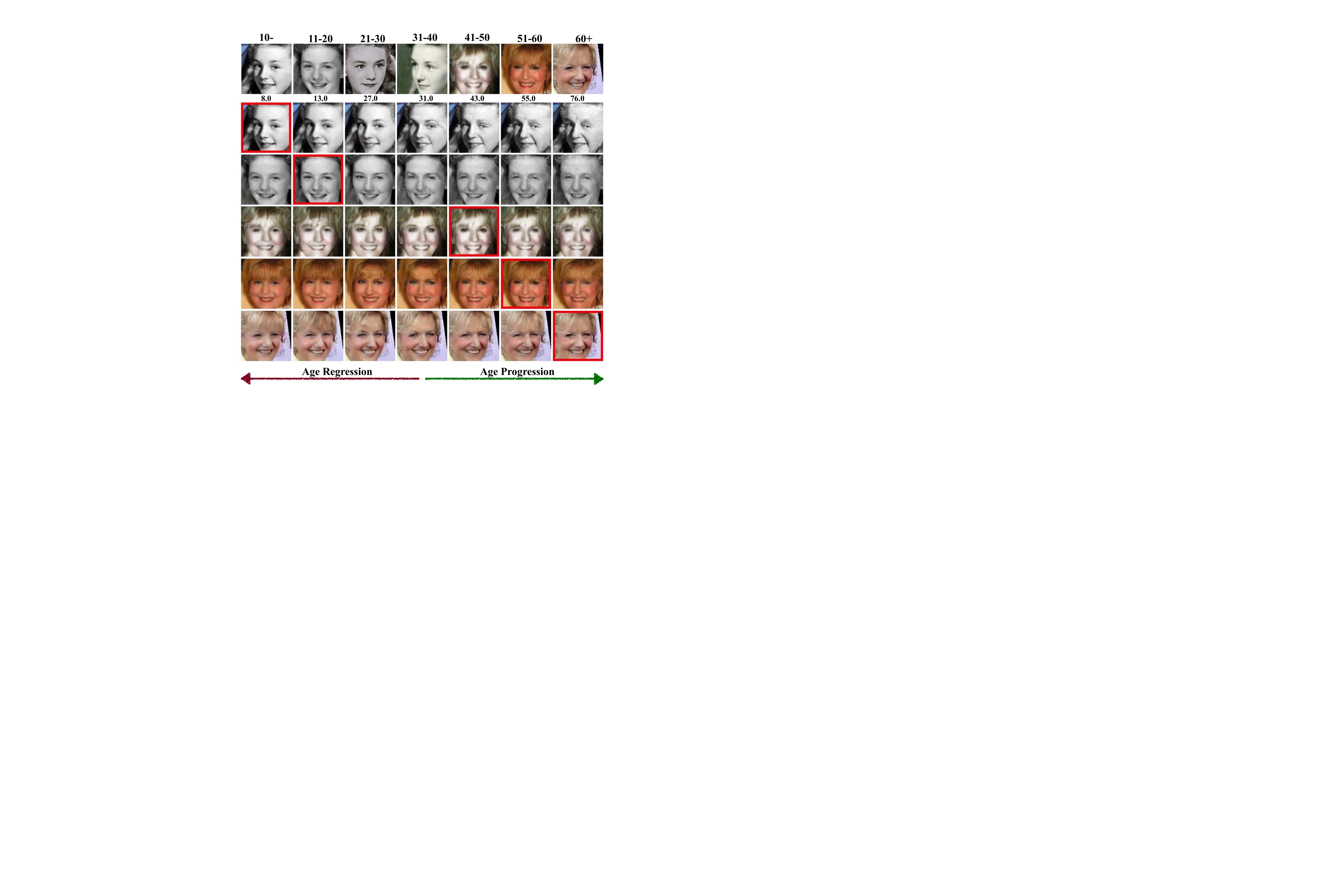}
    \caption{Sample results by our \methodname. First row: the real faces of the same person at different ages with estimated age labels underneath. Remaining rows: the synthesized faces when given input faces in the red boxes.}\vspace{-5mm}
    \label{fig:example}
\end{figure}

Face recognition has been a hot research topic in computer vision for many years. Recently, deep-learning-based methods achieve excellent performance, even surpassing humans in several scenarios, by empowering the face recognition models with deep neural networks~\cite{he2016deep,krizhevsky2017imagenet,simonyan2014very}. The traditional wisdom is to utilize the margin-based metrics to increase the intra-class compactness and train the models with a massive amount of data to improve face recognition performance~\cite{wen2016discriminative}. 

Despite the remarkable success of general face recognition (GFR), \emph{how to minimize the effects of age variation} is a lingering challenge for current face recognition systems to correctly identify faces in many practical applications such as finding lost children. It is therefore of great significance to achieve face recognition without age variation, \ie, age-invariant face recognition or AIFR. However, AIFR remains extremely challenging in the following three aspects. First, when the age gap becomes large in cross-age face recognition, age variation can largely affect the facial appearance, compromising the face recognition performance. Second, face age synthesis (FAS) is a complex process involving face aging/rejuvenation~(\emph{a.k.a} age progression/regression) since the facial appearance drastically changes over a long time and differs from person to person. Last, it is infeasible to obtain a large paired face dataset to train a model in rendering faces with natural effects while preserving identities.

To overcome these issues, current methods for AIFR can be roughly divided into two categories: generative and discriminative models. Give a face image, the generative models~\cite{geng2007automatic,lanitis2002toward,park2010age} aim to transform the faces of different ages into the same age group in order to assist the face recognition. Recently, generative adversarial networks~(GANs)~\cite{goodfellow2014generative} have been successfully used to enhance the image quality of synthesized faces~\cite{li2019age,liu2019attribute,wang2018face,yang2018learning,zhang2017age}; they typically use the one-hot encoding to specify the target age group. However, the one-hot encoding represents the age group-level face transformation, ignoring the identity-level personalized patterns and leading to unexpected artifacts. As a result, the performance of AIFR cannot be significantly improved due to the unpleasing synthesized faces and unexpected changes in identity. On the other hand, the discriminative models~\cite{deb2019finding,wang2018orthogonal} focus on extracting age-invariant features by disentangling the identity-related information from the mixed information so that only the identity-related information is expected for the face recognition systems. Although achieving promising performance in AIFR, they cannot provide users,  for example policemen, with visual results as the generative methods to further verify the identities, which can compromise the model interpretability in the decision-making processes of many practical applications.

To further improve the image quality for generative models and provide the model interpretability for discriminative models, we propose a unified, multi-task learning framework to simultaneously achieve AIFR and FAS, termed \methodname, which can enjoy the best of both worlds; \ie learning age-invariant identity-related representation while achieving pleasing face synthesis. More specifically, we first decompose the mixed high-level features into two uncorrelated components---identity- and age-related features---through an attention mechanism. We then decorrelate these two components in a multi-task learning framework, in which an age estimation task is to extract age-related features while a face recognition task is to extract identity-related feature; in addition, a continuous cross-age discriminator with a gradient reversal layer~\cite{ganin2016domain} further encourages the identity-related age-invariant features.
Moreover, we propose an identity conditional module to achieve identity-level transformation patterns for FAS, with a weight-sharing strategy to improve the age smoothness of synthesized faces; \ie, the faces are aged smoothly.
Extensive experiments demonstrate superior performance over existing state-of-the-art methods for AIFR and FAS, and competitive performance for face recognition in the wild. Fig.~\ref{fig:example} presents an example of age progression/regression of the same person from our \methodname,  showing that our framework can synthesize photorealistic faces while preserving identity. 

Our contributions are summarized as follows. \emph{First}, we propose a unified, multi-task learning framework to jointly handle AIFR and FAS, which can learn age-invariant identity-related representation while achieving pleasing face synthesis. \emph{Second}, we propose an attention-based feature decomposition to separate the age- and identity-related feature on high-level feature maps, which can constrain the decomposition process in contrast to the previous unconstrained decomposition on feature vectors. Age estimation and face recognition tasks are incorporated to supervise the decomposition process in conjunction with a continuous domain adaption. \emph{Third}, compared to previous one-hot encoding achieving age group-level face transformation, we propose a novel identity conditional module to achieve identity-level face transformation, with a weight-sharing strategy to improve the age smoothness of synthesized faces. 
\emph{Fourth}, extensive experiments demonstrate the effectiveness of the proposed framework for AIFR and FAS on five benchmark datasets, and competitive performance on two popular GFR datasets. \emph{Last}, we collect and release a large cross-age dataset of millions of faces with age and gender annotations, which can advance the development of the AIFR and FAS. In addition, it is expected to be useful for other face-related research tasks; \eg, pretraining for face age estimation. 

\begin{figure*}[ht!]
    \centering
    \includegraphics[width=0.8\linewidth]{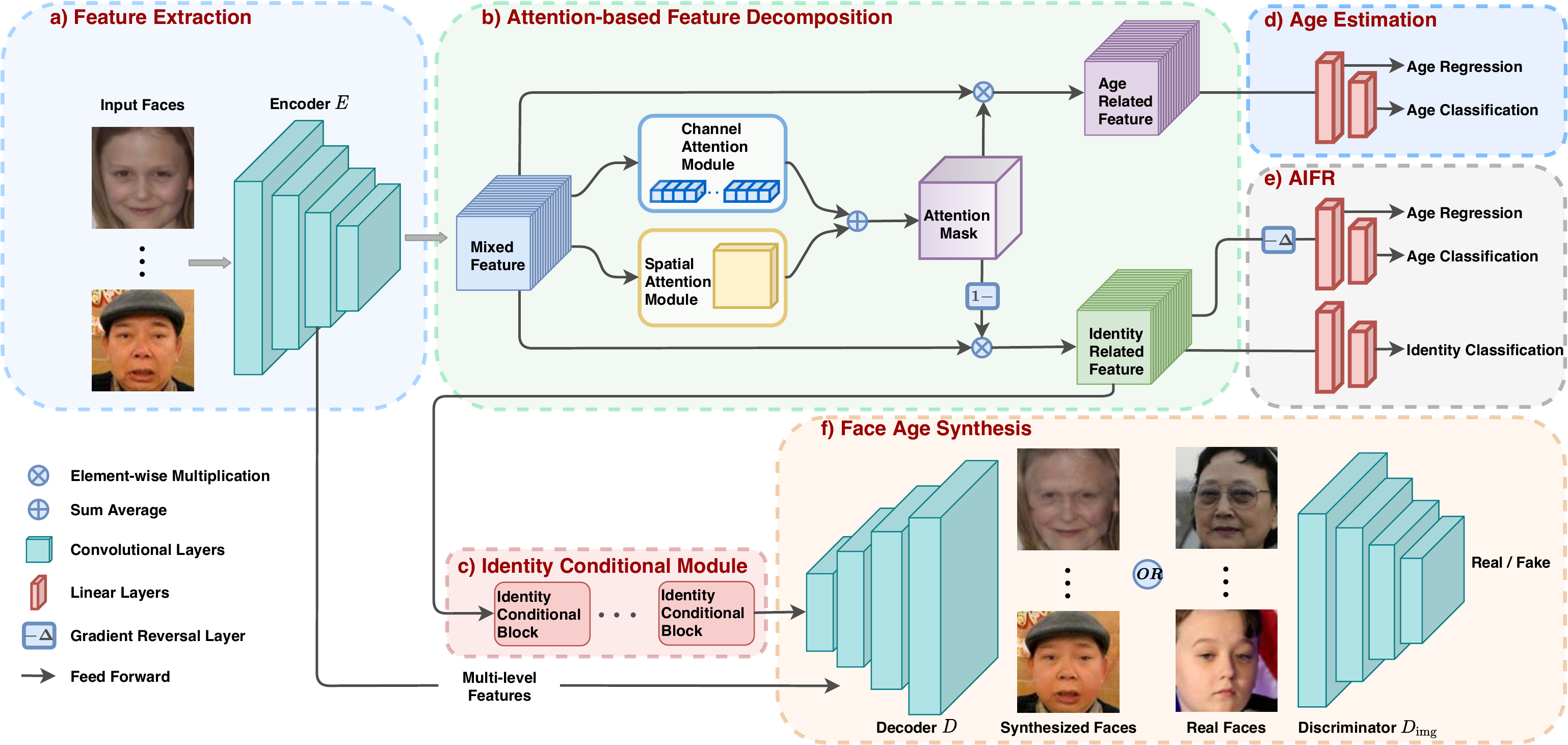}
    \caption{An overview of the proposed \methodname including two tasks. AIFR: The encoder $E$ first extracts the mixed feature maps from input faces, which are then decomposed into two disjoint identity- and age-related feature maps by the multi-task training and continuous domain adaption. FAS: The decoder $D$ produces synthesized faces through identity conditional module based on multi-level features; the PatchDiscriminator $D_{\mathrm{img}}$ penalizes the framework for better visual quality.
    }
    \label{fig:framework}
\end{figure*}
%!tex root=./cvpr.tex
\section{Related Work}
\label{related_work}

\noindent\textbf{Age-invariant face recognition (AIFR).}\quad Prior studies usually tackle age variation by disentangling age-invariant features from mixed features. For example, \cite{gong2013hidden} adopted the hidden factor analysis~(HFA) to factorize the mixed features and then reduce the age variation in identity-related features. \cite{wen2016latent} extended HFA~\cite{gong2013hidden} into a deep learning framework with the latent factor guided convolutional neural network~(LF-CNN). At the same time,~\cite{zheng2017age} introduced an age estimation task to guide the AIFR. Most recently, CNNs-based discriminative methods have achieved promising results for AIFR. OE-CNN~\cite{wang2018orthogonal} adapted a modified softmax loss~\cite{liu2017sphereface} for AIFR by decomposing the facial embeddings into two orthogonal components such that the identity- and age-related features are represented as the angular and radial direction, respectively. Similarly, DAL~\cite{wang2019decorrelated} achieved the feature decomposition in an adversarial manner under the assumption that the two components are uncorrelated. 

The work related to ours is~\cite{zhao2019look}, in which a cGANs-based model, with cross-age domain adversarial training extracting age-invariant representations, is adopted to achieve the two tasks simultaneously. However, it generates oversmoothed faces with subtle changes. Different from~\cite{zhao2019look}, our framework has following advantages: 1) our feature decomposition is done on feature maps through an attention mechanism; 2) a continuous domain adaption with gradient reversal layer is used to learn age-invariant identity-related representation; and 3) identity conditional module can achieve identity-level face synthesis and improve the age smoothness of synthesized faces.

\noindent\textbf{Face age synthesis (FAS).}\quad Existing methods for FAS can be roughly divided into physical model-, and prototype- and deep generative model-based methods. Physical model-based methods~\cite{ramanathan2006modeling,ramanathan2008modeling,suo2012concatenational} mechanically model the changes of appearance over time, but they are computationally expensive and require massive paired images of the same person with a long time. Prototype-based methods~\cite{kemelmacher2014illumination,rowland1995manipulating} achieve face aging/rejuvenation using the average of faces in each age group, hence the identity cannot be well preserved. The deep generative model-based methods~\cite{Duong_2017_ICCV,wang2016recurrent} exploit the deep neural network for this task. For example, recurrent face aging~(RFA)~\cite{wang2016recurrent} used a recurrent neural network to model the intermediate transition states of age progression/regression, traversing on which a smooth face aging process can be achieved. Inspired by the powerful capability of generative adversarial networks~(GANs)~\cite{goodfellow2014generative}, especially conditional GANs~(cGANs)~\cite{mirza2014conditional}, in generating high-quality images, many recent studies~\cite{huang2020pfa,zhang2017age,wang2018face,yang2018learning} resort to them to improve the visual quality of synthesized faces and train models with unpaired age data. For example, \cite{zhang2017age} used a conditional adversarial autoencoder~(CAAE) to achieve both age progression/regression by traversing on a low-dimensional face manifold. \cite{wang2018face} introduced the perceptual loss to preserve the identities during face aging/rejuvenation. \cite{yang2018learning} designed a discriminator with the pyramid architecture to enhance the aging details. 

However, these methods mainly aim at improving the visual quality of generated faces, and hardly improve the performance of AIFR due to the artifacts resulting from group-level face transformation, and the unexpected change in identity. Our method differs in the following aspects: 1) the proposed \methodname achieves AIFR and FAS simultaneously to enhance the visual quality with identity-related information from AIFR; 2) the proposed identity conditional module~(ICM) achieves an identity-level face age synthesis in contrast to the previous group-level face age synthesis; and 3) a weight-sharing strategy in ICM can improve the age smoothness of synthesized faces.
%!tex root=./cvpr.tex
\section{Methodology}\label{sec:method}

Fig.~\ref{fig:framework} presents the architecture of the proposed \methodname, which will be detailed in the following subsections. 

\subsection{Attention-based Feature Decomposition}

As the faces change a lot over time, the critical problem of AIFR is that the age variation usually introduces the increasing intra-class distances. As a result, it is challenging to correctly recognize two faces of the same person with a large gap, since the mixed facial representations are severely entangled with unrelated information such as facial shape and texture changes. Recently, Wang~\etal~design a linear factorization module to decompose the feature vectors into these two unrelated components~\cite{wang2019decorrelated}. Formally, given the feature vector $\vct{x}\in\mathbb{R}^d$ extracted from an input image $\mat{I}\in \mathbb{R}^{3\times H\times W}$, their linear factorization module is defined as~\cite{wang2019decorrelated}:
\begin{align}\label{eq_linear_decomposition}
    \mat{x} = \mat{x}_{\mathrm{age}} + \mat{x}_{\mathrm{id}},
\end{align}
where $\mat{x}_{\mathrm{age}}$ and $\mat{x}_{\mathrm{id}}$  denote the age- and identity-related components, respectively. This decomposition is implemented through a residual mapping. However, it has the following drawbacks: 1) this decomposition performs on one-dimensional feature vector, the resultant identity-related component lacks spatial information of face, not suitable for FAS; and 2) this decomposition is unconstrained, which may lead to unstable training. 

To address these drawbacks, we instead propose to decompose the mixed feature-maps at a high-level semantic space through an attention mechanism, termed attention-based feature decomposition or AFD. The main reason is that manipulating on the feature vectors is more complicated than on the feature maps since the aging/rejuvenation effects, such as beards and wrinkles, appear in the semantic feature space but lose in the one-dimensional features.  Formally, we use a ResNet-like backbone as encoder $E$ to extract mixed feature maps  $\mat{X} \in \mathbb{R}^{C\times H'\times W'}$ from an input image $\mat{I}$, \ie $\mat{X} = E(\mat{I})$, the AFD can be defined as follows:
\begin{align}
\mat{X} = \underbrace{\mat{X} \circ \sigma(\mat{X})}_{\mat{X}_{\mathrm{age}}} + \underbrace{\mat{X} \circ \big(1 - \sigma(\mat{X})\big)}_{\mat{X}_{\mathrm{id}}},
\end{align}
where $\circ$ denotes element-wise multiplication and $\sigma$ represents an attention module. In doing so, the age-related information in the feature maps can be separated through the attention module supervised by an age estimation task, and the residual part, regarded as the identity-related information, can be supervised by a face recognition task. As a result, the attention mechanism constrains the decomposition module, better at detecting the age-related features in semantic feature maps. We note that $\mat{X}$ is assumed to only contain the age and identity information as driven by the two corresponding tasks, the remaining information such as background is important for FAS, which is preserved by skip connections from encoder to decoder. Fig.~\ref{fig:framework}(b) details the proposed AFD. 

In this paper, we adopt the sum average of channel attention~(CA)~\cite{hu2018squeeze} and spatial attention~(SA)~\cite{woo2018cbam} to highlight age-related information at both channel and spatial levels. Note that the outputs of these two attentions have different sizes, we first stretch each of them to the original input size and then average them. Different attention modules such as CA, SA, and CBAM~\cite{woo2018cbam} are also investigated in Sec.~\ref{sec:exp}.

\subsection{Identity Conditional Module}

\begin{figure}[ht!]
    \centering
    \includegraphics[width=0.8\linewidth]{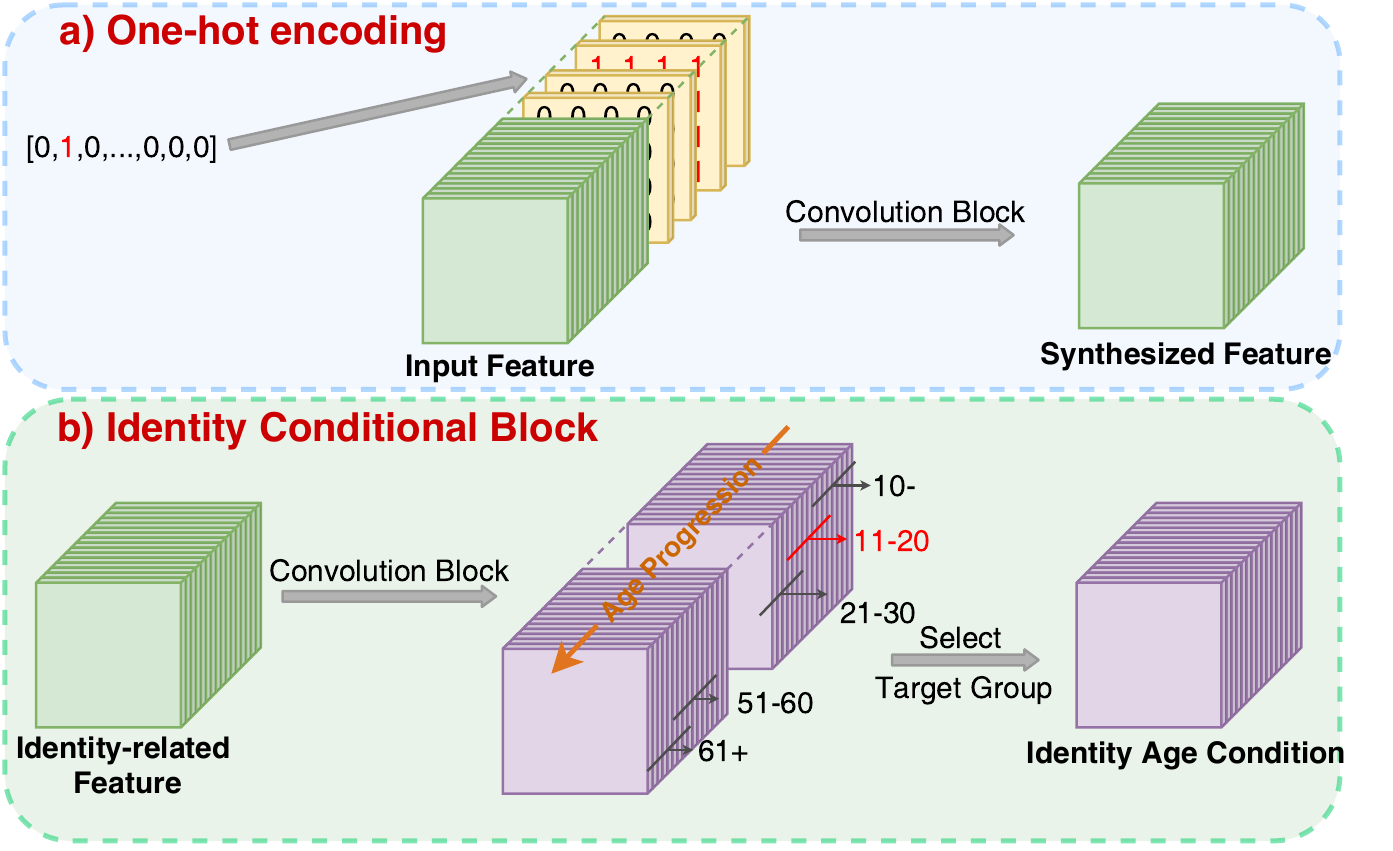}
    \caption{Comparison between one-hot encoding and ICB. 
    }
    \label{fig:condition}
\end{figure}

The mainstream face aging studies~\cite{li2019age,liu2019attribute,wang2018face,yang2018learning,zhang2017age} usually split the ages into several non-overlapping age groups, since the changes over time are minor with a small age gap. These methods typically use one-hot encoding to specify the target age group to control the aging/rejuvenation process~\cite{li2019age,wang2018face,zhang2017age} as illustrated in Fig.~\ref{fig:condition}(a). Consequently, a group-level aging/rejuvenation pattern, such as people having a beard when they are 30 years old, is learned for each age group due to the use of one-hot age condition. Its drawbacks are twofold: 1) one-hot encoding represents the age group-level aging/rejuvenation pattern, ignoring identity-level personalized pattern, particularly for different genders and races; and 2) one-hot encoding may not ensure the age smoothness of synthesized faces.

To address these issues raised by one-hot encoding, we propose an identity conditional block (ICB) to achieve identity-level aging/rejuvenation pattern, with a weight-sharing strategy to improve the age smoothness of synthesized faces.
Specifically, the proposed ICB takes the identity-related feature from AFD as input to learn an identity-level aging/rejuvenation pattern. Next,  we propose a weights-sharing strategy to improve the age smoothness of synthesized faces so that some convolutional filters are shared across adjacent age groups as shown in Fig.~\ref{fig:condition}(b).
The rationale behind this idea is that faces are gradually changed over time, where the shared filters can learn some common aging/rejuvenation patterns between adjacent age groups. Note that $\mat{X}_{\mathrm{id}}$ is reduced from $512$ to $128$ using $1\times 1$ convolutions to reduce the computational cost. In this paper, a hyper-parameter $s$ to control how many filters are shared for two adjacent age groups, which is empirically set to $1/8$; \ie, the adjacent two age groups share 16 filters. We stack ICBs to form an identity conditional module (ICM). 

\subsection{Multi-task Learning Framework}

This section presents our \methodname including AIFR and FAS.

\noindent\textbf{Age-invariant face recognition (AIFR) task.}\quad To encourage AFD to robustly decompose features, we use an age estimation task and a face recognition task to supervise the feature decomposition. Specifically, $\mat{X}_{\mathrm{age}}$ draws the age variations by an age estimation task while $\mat{X}_{\mathrm{id}}$ encodes the identity-related information. To be clear, we include an age estimation network $A$ with two linear layers of 512 and 101 neurons to achieve age regression similar to deep expectation~(DEX)~\cite{rothe2015dex} that learns the age distribution by computing a softmax expected value. Another linear layer $\boldsymbol{W} \in \mathbb{R}^{101 \times N}$ is appended at the last for age classification, regularizing the learned distribution; here, $N$ is the number of age groups. The loss function to optimize age estimation can be defined as:
\begin{align}
    \ell_{\textsc{ae}}(\mat{X}_{\mathrm{age}}) =\mathbb{E}_{\mat{I}}\big[&\ell_{\textsc{mse}}\left(\mathrm{DEX}({A}(\mat{X}_{\mathrm{age}})), y_\mathrm{age}\right) \notag\\
    &+ \ell_{\textsc{ce}}\left({A}(\mat{X}_{\mathrm{age}})\mat{W}, c_{\mathrm{age}}\right)\big],
\end{align}
where $y_\mathrm{age}$, $c_{\mathrm{age}}$, $\ell_{\textsc{mse}}$, and $\ell_{\textsc{ce}}$ are the ground truth age,  ground truth age group, and mean squared error (MSE) for age regression, cross-entropy (CE) loss for age group classification, respectively. 

Next, we leverage one linear layer $L$ of 512 neurons to extract the feature vectors, and use the CosFace loss to supervise the learning of $\mat{X}_{\mathrm{id}}$ for identity classification. We also introduce a cross-age domain adversarial learning that encourages $\mat{X}_{\mathrm{id}}$ to be age-invariant through a continuous domain adaption~\cite{wang2020continuously} with a gradient reversal layer~(GRL)~\cite{ganin2016domain}.
The final loss for AIFR is formulated as: 
\begin{align}
\mathcal{L}^{\textsc{aifr}} =& 
\ell_{\textsc{cosface}}(L(\mat{X}_{\mathrm{id}}), y_\mathrm{id})  \\
&+\lambda^{\textsc{aifr}}_{\mathrm{age}}\mathcal{L}_{\textsc{ae}}(\mat{X}_{\mathrm{age}})+
\lambda_{\mathrm{id}}^{\textsc{aifr}} \mathcal{L}_{\textsc{ae}}(\mathrm{GRL}(\mat{X}_{\mathrm{id}})),\notag
\end{align}
where the first term is the CosFace loss, the second term is the age estimation loss, and the last term is the domain adaption loss, $y_\mathrm{id}$ is the identity label, and $\lambda_{\mathrm{*}}$ controls the balance of different loss terms. Note that the second and third terms use the same network structure, but have different inputs and are trained independently. The activation functions and batch normalizations are ignored for simplicity, and our face recognition model is designed strictly following the setting in~\cite{deng2019arcface} except the AFD.

\noindent\textbf{Face age synthesis~(FAS) task.}\quad 
Fig.~\ref{fig:framework}(f) demonstrates the FAS process of our proposed method. In detail, the identity-level age condition is derived from the discriminative facial representations $\mat{X}_{\mathrm{id}}$ by applying an identity conditional module~(ICM) with a series of ICBs. Then, the decoder $D$ reconstructs the progressed/regressed faces from the multi-level high-resolution features extracted from the encoder $E$, under the control of the learned identity-level age condition. Formally, the process of rendering input face $\mat{I}$ to $\mat{I}_t$ with target age group $t$ can be written as:
\begin{align}
    \mat{I}_t=D\big(\{E_l(\mat{I})\}_{l=1}^{3}, \mathrm{ICM}(\mat{X}_{\mathrm{id}}, t)\big),
\end{align}
where $l$ denotes the index of different levels of high-resolution features extracted from different layers of the encoder $E$.

To facilitate the visual quality of generated faces, the FAS task is trained using GANs framework. 
In this paper, we adopt the PatchDiscriminator from~\cite{isola2017image} as our discriminator $D_\mathrm{img}$ to emphasize the local-patch of generated and real images. Furthermore, the least-squares GANs~\cite{mao2017least} are employed to optimize the GANs framework for an improved quality of generated images and stable training process, which can be formulated as follows:
\begin{align}
    \mathcal{L}_{\mathrm{adv}}^{\textsc{fas}}=\frac{1}{2} \mathbb{E}_{\mat{I}}\left[D_\mathrm{img}([\mat{I}_t; \mat{C}_t])-1\right]^{2},
\end{align}
where $\mat{C}_t$ is the one-hot encoding used in traditional cGANs framework for aligning the age condition, and $[;]$ denotes the matrix concatenation along channel dimension. To preserve the identities of input faces and improve the age accuracy, we leverage the encoder $E$ and AFD to supervise the FAS task. Consequently, we can achieve both face aging and rejuvenation in a holistic, end-to-end manner, as illustrated in Fig.~\ref{fig:framework}. This process can be formulated as follows:
\begin{align}
    \mat{X}_{\mathrm{age}}^t, \mat{X}_{\mathrm{id}}^t &= \mathrm{AFD}\big(E(\mat{I}_t)\big) \\
    \mathcal{L}_{\mathrm{age}}^{\textsc{fas}} 
    &= \ell_{\textsc{ce}}\big(\mat{X}_{\mathrm{age}}^t, t\big), \\
    \mathcal{L}_{\mathrm{id}}^{\textsc{fas}} &= \mathbb{E}_{\mat{X}_{s}}\left\|\mat{X}_{\mathrm{id}}^t-\mat{X}_{\mathrm{id}}\right\|^2_F,
\end{align}
where $\|\cdot\|_F$ represents the Frobenius norm.

The final loss to optimize this task can be written as:
\begin{align}
    \mathcal{L}^{\textsc{fas}} = \lambda_{\mathrm{adv}}^{\textsc{fas}} \mathcal{L}_{\mathrm{adv}}^{\textsc{fas}} + \lambda_{\mathrm{id}}^{\textsc{fas}} \mathcal{L}_{\mathrm{id}}^{\textsc{fas}} + \lambda_{\mathrm{age}}^{\textsc{fas}} \mathcal{L}_{\mathrm{age}}^{\textsc{fas}},
\end{align}
where $\lambda_{*}^{\textsc{fas}}$ controls the importance of different loss terms of FAS task. The loss function to optimize the discriminator $D_\mathrm{img}$ in the context of least-squares GANs is defined as:
\begin{align}
    \mathcal{L}_{D_{\mathrm{img}}}^{\textsc{fas}} =& \frac{1}{2} \mathbb{E}_{\mat{I}} \left[D_{\mathrm{img}}\big([\mat{I};\mat{C}]\big)-1\right]^{2} \notag\\
    &+
    \frac{1}{2} \mathbb{E}_{\mat{I}_{t}} \left[D_{\mathrm{img}}\big([\mat{I}_{t};\mat{C}_t]\big)\right]^{2}.
\end{align}

At the testing stage, the only difference from existing FAS methods is that our method needs to specify the corresponding group of filters. Consequently, our method enjoys the advantages similar to~\cite{he2019s2gan} that the computational cost can be significantly reduced by only encoding input faces once, instead of $N$ times in previous works~\cite{li2019age,liu2019attribute,wang2018face,yang2018learning,zhang2017age}, where $N$ is the number of age groups.

\noindent\textbf{Optimization and inference.}\quad In our \methodname, the AIFR learns the discriminative facial representations and age estimation while the FAS produces the visual results which can boost the model interpretability for AIFR. Therefore, both two tasks can be jointly accomplished by optimizing these two tasks in a GAN-like manner; they mutually leverage each other to boost themselves. In other words, the AIFR encourages FAS to render faces to preserve its identity while FAS can facilitate the extraction of the identity-related feature and boost the model interpretability for AIFR. Consequently, we alternately train these two tasks in a unified, multi-task, end-to-end framework.
%!tex root=./cvpr.tex
\section{Experiments}\label{sec:exp}

\subsection{Implementation Details}

\begin{figure}[t]
    \centering
    \includegraphics[width=0.8\linewidth]{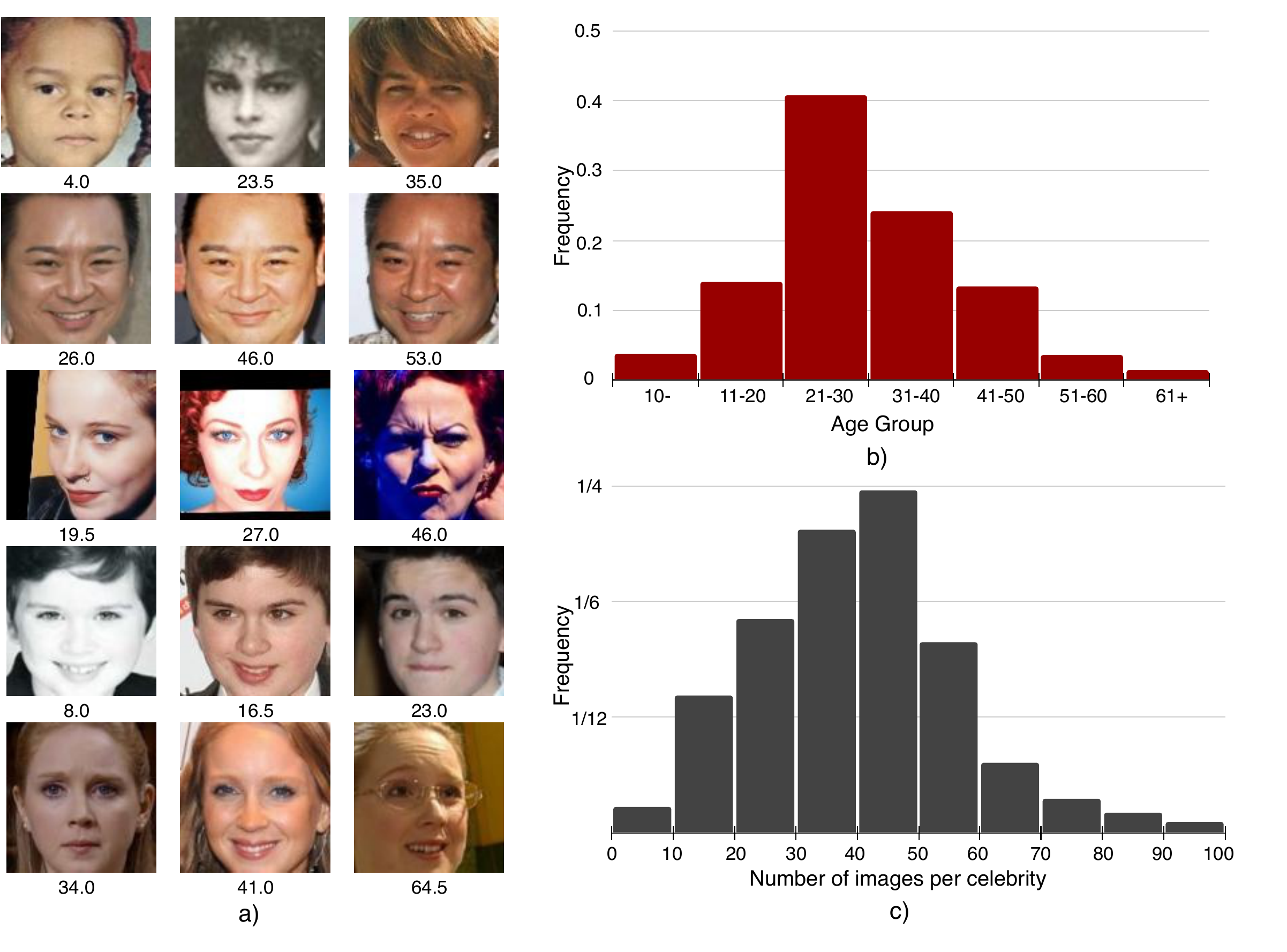}
    \caption{
        Sample faces (a) and dataset statistics (b and c) on SCAF.
    }
    \label{fig:dataset}
    \vspace{-4mm}
\end{figure}

\noindent\textbf{Data collection.}\quad Current research on AIFR lacks a large-scale face dataset of millions of face images with a large age gap. To advance the development of AIFR and FAS, we create and release a new large cross-age face dataset~(LCAF) with 1.7M faces from cross-age celebrities. Specifically, we first use the public Azure Facial API~\cite{azure} to estimate the ages and genders of faces from the clean MS-Celeb-1M dataset provided by~\cite{deng2019arcface}. Then, we randomly select faces from a total of 5M faces to check whether the faces are correctly labeled, and try our best to manually correct them if any apparent mistakes; we mainly focus on the young ages under 20 that are often mislabeled by the API~\cite{azure}. Finally, a large-scale balanced age dataset is constructed by balancing both age and gender. We further build a subset of cross-age face dataset~(SCAF) containing about 0.5M images from 12K individuals following~\cite{wang2019decorrelated,wang2018orthogonal} for fair comparisons. We note that the training (LCAF) and testing data may have very little, or even no identities overlapping as \cite{deng2019arcface} already removed 500+ identities from their clean MS-Celeb-1M dataset by checking the similarity of faces between training and testing data. Following the mainstream literature~\cite{he2019s2gan,lihierarchical,li2019age,liu2019attribute,yang2018learning} with the time span of 10 years for each age group, the ages in this paper are divided into seven non-overlapping groups; \ie, 10-, 11-20, 21-30, 31-40, 41-50, 51-60, and 61+. Note that it is a much more challenging problem to perform FAS on seven groups than on four groups in previous works. Fig.~\ref{fig:dataset} presents example images and dataset statistics of SCAF. 

\noindent\textbf{Training details.}\quad We adopted ResNet-50 similar to~\cite{deng2019arcface} as the encoder $E$. In the decoder $D$, the identity age condition is bilinearly upsampled and processed with multi-level high-resolution features extracted from $E$ by two ResBlocks~\cite{he2016deep}, each of which follows the instance normalization~\cite{ulyanov2016instance} and ReLU activation, the synthesized faces of size $112 \times 112$ were produced by one $1\times 1$ convolutional layer. There are four ICBs in ICM. In the discriminator $D_\mathrm{img}$, six convolutional layers with strides of 2, 2, 2, 2, 1, 1 follow the spectral normalization~\cite{miyato2018spectral} and leaky ReLU activation except the last one, outputting $7\times 7$ confidence map. The AIFR is optimized by SGD with an initial learning rate of $0.1$ and momentum of 0.9 while the ICM, decoder $D$, and $D_\mathrm{img}$ are trained by Adam with a fixed learning rate of $10^{-4}$, $\beta_1$ of 0.9 and $\beta_2$ of 0.99 for the face age synthesis. We trained all models with a batch size of 512 on 8 NVIDIA GTX 2080Ti GPUs, 110K iterations for LCAF and 36K iterations for SCAF. The learning rate of AIFR was warmed up linearly from 0 to 0.1, reduced by a factor of 0.1, at iterations 5K, 70K, and 90K on LCAF and 1K, 20K, 23K on SCAF, respectively. The hyper-parameters in the loss functions were empirically set as follows: $\lambda^{\textsc{aifr}}_{\mathrm{age}}$ was $0.001$, $\lambda^{\textsc{aifr}}_{\mathrm{id}}$ was $0.002$, $\lambda_{\mathrm{adv}}^{\textsc{fas}}$ was $75$, $\lambda_{\mathrm{id}}^{\textsc{fas}}$ was $0.002$, and $\lambda_{\mathrm{age}}^{\textsc{fas}}$ was $10$. The multiplicative margin and scale factor of CosFace loss~\cite{wang2018cosface} were set to $0.35$ and $64$, respectively. All images were aligned to $112\times 112$, with five facial landmarks detected by MTCNN~\cite{zhang2016joint}, and linearly normalized to $[-1, 1]$.

\subsection{Evaluation on AIFR}

\begin{table*}[t]
    \centering
    \scriptsize
    \begin{subtable}[b]{0.2\textwidth}
        \begin{tabular}{lc}
            \toprule
            Method                                      & Acc~(\%) \\
            \midrule
            RJIVE~\cite{sagonas2017recovering}          & 55.20      \\
            VGG Face~\cite{parkhi2015deep}              & 89.89      \\
            Center Loss~\cite{wen2016discriminative}    & 93.72      \\
            SphereFace~\cite{liu2017sphereface}         & 91.70      \\
            CosFace~\cite{wang2018cosface}              & 94.56      \\
            ArcFace~\cite{deng2019arcface}              & 95.15      \\
            DAAE~\cite{lihierarchical}                  & 95.30      \\
            \midrule
            \methodname(\textbf{ours})                               & \textbf{96.23}        \\
            \bottomrule
            \end{tabular}
            \caption{\bd{AgeDB-30}}\label{tab:agedb}
    \end{subtable}
    \hspace{0.05\textwidth}
    \begin{subtable}[b]{0.2\textwidth}
    \begin{tabular}{lc}
        \toprule
        Method        & Acc~(\%) \\
        \midrule
        HUMAN-Individual  & 82.32      \\
        HUMAN-Fusion      & 86.50      \\
        \midrule
        Center Loss~\cite{wen2016discriminative}   & 85.48      \\
        SphereFace~\cite{liu2017sphereface}        & 90.30      \\
        VGGFace2~\cite{cao2018vggface2}            & 90.57     \\
        ArcFace~\cite{deng2019arcface}             & 95.45     \\
        \midrule
        \methodname(\textbf{ours}) & \textbf{95.62}      \\
        \bottomrule
    \end{tabular}
    \caption{\bd{CALFW}}\label{tab:calfw}
    \end{subtable}
    \hspace{0.05\textwidth}
    \begin{subtable}[b]{0.2\textwidth}
    \begin{tabular}{lc}
        \toprule
        Method        & Acc~(\%) \\
        \midrule
        HFA~\cite{gong2013hidden}                 & 84.40      \\
        CARC~\cite{chen2015face}                  & 87.60      \\
        VGGFace~\cite{parkhi2015deep}             & 96.00      \\
        Center Loss~\cite{wen2016discriminative}  & 97.48     \\
        LF-CNN~\cite{wen2016latent}               & 98.50      \\
        Marginal Loss~\cite{deng2017marginal}     & 98.95      \\
        OE-CNN~\cite{wang2018orthogonal}          & 99.20      \\
        AIM~\cite{zhao2019look}                   & 99.38      \\
        DAL~\cite{wang2019decorrelated}           & 99.40      \\
        \midrule
        \methodname(\textbf{ours}) & \textbf{99.55}      \\
        \bottomrule
    \end{tabular}
    \caption{\bd{CACD-VS}}\label{tab:cacdvs}
    \end{subtable}
    \hspace{0.05\textwidth}
    \begin{subtable}[b]{0.2\textwidth}
    \begin{tabular}{lc}
        \toprule
        Method        & Rank-1~(\%) \\
        \midrule
        Park~\etal~\cite{park2010age}                   & 37.40      \\
        Li~\etal~\cite{li2011discriminative}            & 47.50      \\
        HFA~\cite{gong2013hidden}                       & 69.00      \\
        MEFA~\cite{gong2015maximum}                     & 76.20      \\
        CAN~\cite{xu2017age}                            & 86.50      \\
        LF-CNN~\cite{wen2016latent}                     & 88.10      \\
        AIM~\cite{zhao2019look}                         & 93.20      \\
        DAL~\cite{wang2019decorrelated}                 & 94.50      \\
        \midrule
        \methodname(\textbf{ours}) & \textbf{94.78}      \\
        \bottomrule
    \end{tabular}
    \caption{\bd{FG-NET~{(\footnotesize leave-one-out)}}}\label{tab:fgnet}
    \end{subtable}\\
    \vspace{4mm}
    \begin{subtable}[b]{0.2\textwidth}
    \begin{tabular}{lc}
        \toprule
        Method        & Rank-1~(\%) \\
        \midrule
        FUDAN-CS\_SDS~\cite{wang2017multi}          & 25.56      \\
        SphereFace~\cite{liu2017sphereface}            & 47.55      \\
        TNVP~\cite{Duong_2017_ICCV}           & 47.72      \\
        OE-CNN~\cite{wang2018orthogonal}           & 52.67      \\
        DAL~\cite{wang2019decorrelated}           & \textbf{57.92}      \\
        \midrule
        \methodname(\textbf{ours}) & 57.18      \\
        \bottomrule
    \end{tabular}
    \caption{\bd{FG-NET~(MF1)}}\label{tab:fgnet_mf1}
    \end{subtable}
    \hspace{0.04\textwidth}
    \begin{subtable}[b]{0.405\textwidth}
    \begin{tabular}{lccccc}
        \toprule
        Model & AgeDB-30 & CALFW & CACD-VS & FG-NET \\ 
        \midrule
        Baseline & 95.52 & 94.27 & 99.12 & 93.64 \\
        \quad+Age & 95.32 & 94.35 & 99.15 & 93.88 \\ 
        \quad+AFD~(CA) & 95.63 & 94.50 & 99.32 & 94.05 \\ 
        \quad+AFD~(SA) & 95.85 & 94.43 & 99.25 & 94.38 \\ 
        \quad+AFD~(CBAM) & 96.08 & 94.32 & 99.18 & 94.36 \\ 
        \quad+AFD & 95.90 & 94.48 & 99.30 & 94.58 \\ 
        \midrule
        \methodname(\textbf{ours}) & \textbf{96.23} & \textbf{94.72} & \textbf{99.38} & \textbf{94.78}\\ 
        \bottomrule
    \end{tabular}
    \caption{\bd{Ablation Study}}\label{tab:ablation_study}
    \end{subtable}
    \hspace{0.04\textwidth}
    \begin{subtable}[b]{0.28\textwidth}
    \begin{tabular}{lcc}
        \toprule
        Method        & LFW   & MF1-Facescrub \\
        \midrule
        SphereFace~\cite{liu2017sphereface}    & 99.42 & 72.73         \\
        CosFace~\cite{wang2018cosface}         & 99.33 & 77.11         \\
        OE-CNN~\cite{wang2018orthogonal}       & 99.35 & N/A           \\
        DAL~\cite{wang2019decorrelated}        & 99.47 & \textbf{77.58}         \\
        \midrule
        \methodname(\textbf{ours}) & \textbf{99.52} & 77.06         \\ 
        \bottomrule
    \end{tabular}
    \caption{\bd{General Face Recognition}}\label{tab:gfr}
    \end{subtable}
    % main caption
    \caption{
        Experimental results on several benchmark AIFR and GFR datasets with the best results in bold.
    We reported the verification rate~(\%) for AgeDB, CALFW, CACD-VS, and LFW, and the rank-1 identification rate~(\%) for FG-NET and MF1.
    }
    \label{tab:results}
\end{table*}

Next, we evaluate the \methodname on several benchmark cross-age datasets, including CACD-VS~\cite{chen2015face}, CALFW~\cite{zheng2017cross}, AgeDB~\cite{moschoglou2017agedb}, and FG-NET~\cite{fgnet}, to compare with the state-of-the-art methods. Note that MORPH is excluded since the version in~\cite{wang2019decorrelated,wang2018orthogonal,zhao2019look} is prepared for commercial use only.

\noindent\textbf{Result on AgeDB.}\quad AgeDB~\cite{moschoglou2017agedb} contains 16,488 face images of 568 distinct subjects with manually annotated age labels. It provides four protocols for age-invariant face verification protocols under the different age gaps of face pair; \ie 5, 10, 20, and 30 years. Similar to LFW~\cite{huang2008labeled}, this dataset is split into 10 folds for each protocol, where each fold consists of 300 intra-class and 300 inter-class pairs. We strictly follow the protocol of $30$ years to perform the 10-fold cross-validation since the protocol of $30$ years is the most challenging. We use the models trained on SCAF to evaluate the performance on AgeDB for fair comparisons. Table~\ref{tab:agedb} shows the verification accuracy of our models compared against the other state-of-the-art AIFR methods, demonstrating the superior performance of the proposed method.

\noindent\textbf{Result on CALFW.}\quad Cross-age labeled faces in the wild~(CALFW) dataset~\cite{zheng2017cross} is designed for unconstrained face verification with large age gaps, which contains 12,176 face images of 4,025 individuals collected using the same identities in LFW. Similarly, we follow the same protocol as the LFW, where each fold consists of 600 positive and negative pairs. We train the model on LCAF to evaluate our method on this dataset, and the results are shown in Table~\ref{tab:calfw}. Particularly, our method outperforms the recent state-of-the-art AIFR methods by a large margin, establishing a new state-of-the-art on the CALFW.

\noindent\textbf{Result on CACD-VS.}\quad As a public age dataset for AIFR, cross-age celebrity dataset~(CACD) contains 163,446 face images of 2,000 celebrities in the wild, with significant variations in age, illumination, pose, and so on. Since collected by search engine, CACD is noisy with mislabeled and duplicate images. Therefore, a carefully annotated version, CACD verification sub-set or CACD-VS, is constructed for fair comparisons, which also follows the protocol of LFW. Table~\ref{tab:cacdvs} presents the comparison of the proposed method with other state-of-the-arts on CACD-VS~\cite{chen2015face}. Our \methodname surpasses other state-of-the-arts by a large margin, introducing an improvement of 0.15 against the recent one.

\noindent\textbf{Result on FG-NET.}\quad FG-NET~\cite{fgnet} is the most popular and challenging age dataset for AIFR, which consists of 1,002 face images from 82 subjects collected from the wild with huge age variations ranging from child to elder. We strictly follow the evaluation pipeline in~\cite{wang2019decorrelated,wang2018orthogonal}. Specifically, the model is trained on SCAF and tested under the protocols of leave-one-out and MegaFace challenge 1~(MF1). In the leave-one-out protocol, faces are used to match the rest faces, repeating 1,002 times. Table~\ref{tab:fgnet} reports the rank-1 recognition rate. Our method outperforms prior work by a large margin. On the other hand, the MF1 contains additional 1M images as the distractors in the gallery set from 690K different individuals, where models are evaluated under the large and small training set protocols. The small protocol requires the training set less than 0.5M images. The small protocol is strictly followed to evaluate our trained model on FG-NET, and the experimental results are reported in Table~\ref{tab:fgnet_mf1}. Our method achieves competitive performance against other methods since the distractors in MF1 contains a large number of mislabeled probe and gallery face images.

\begin{figure*}[t]
    \centering
    \includegraphics[width=0.83\linewidth]{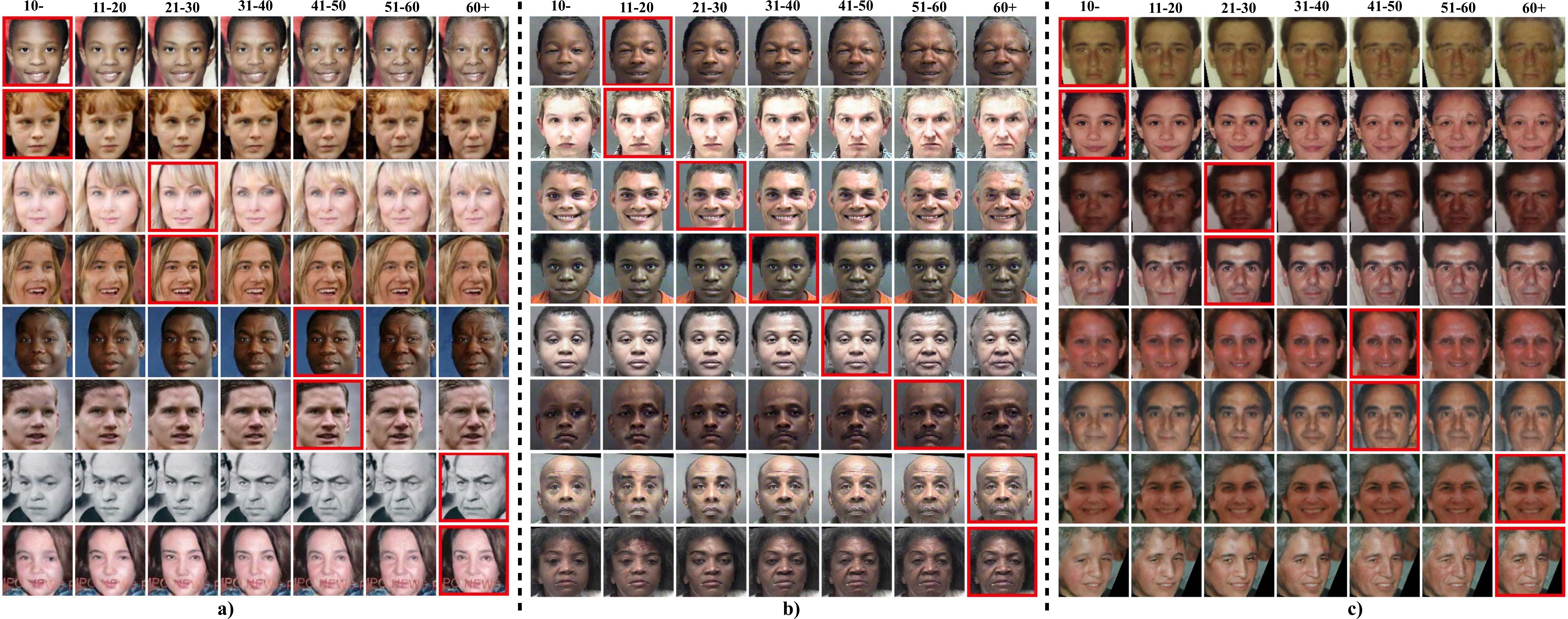}
    \caption{Qualitative results by applying our \methodname trained on SCAF dataset to three external datasets : a) LCAF excluding identities in SCAF; b) MORPH; and c) FG-NET. Red boxes indicate input faces.}
    \label{fig:qualitative_results}
\end{figure*}

\noindent\textbf{Ablation study.}\quad To investigate the efficacy of different modules in~\methodname, we perform ablation studies based on four benchmark datasets for AIFR by considering the following variants of our method:
1) Baseline: we remove all extra components but only the CosFace loss to train the face recognition model. 
2) +Age: this variant is jointly trained under the supervision of both CosFace and age estimation loss, similar to~\cite{wang2019decorrelated,zheng2017age}. 
3) +AFD~(CA), +AFD~(SA), +AFD~(CBAM), +AFD: these four variants utilize the proposed attention-based feature decomposition to highlight the age-related information at different level, by the different attention modules including CA~\cite{hu2018squeeze}, SA~\cite{woo2018cbam}, CBAM~\cite{woo2018cbam}, and the proposed one. 
4) Ours: our proposed~\methodname is trained simultaneously by the AFD and cross-age domain adaption loss. The experimental results are reported in Table~\ref{tab:ablation_study}. 
We note that the verification rate of the baseline model on AgeDB-30 is higher than those of ArcFace and DAAE since the training data (\ie, SCAF) is age-balanced, which is an important feature of our collected dataset. Even though the age estimation task is performed in the face recognition model, it cannot introduce any improvement of AIFR compared to the baseline model. On the other hand, AFD achieves remarkable performance improvement on all cross-age datasets. Nevertheless, as the AFD highlights the age-related information at both channel and spatial levels in parallel, our method achieves consistent performance improvements, demonstrating its effectiveness compared to the single level~(CA and SA) or sequential level~(CBAM). Furthermore, the use of cross-age domain adversarial training leads to an additional performance improvement.

\begin{figure}[ht!]
    \centering
    \includegraphics[width=.8\linewidth]{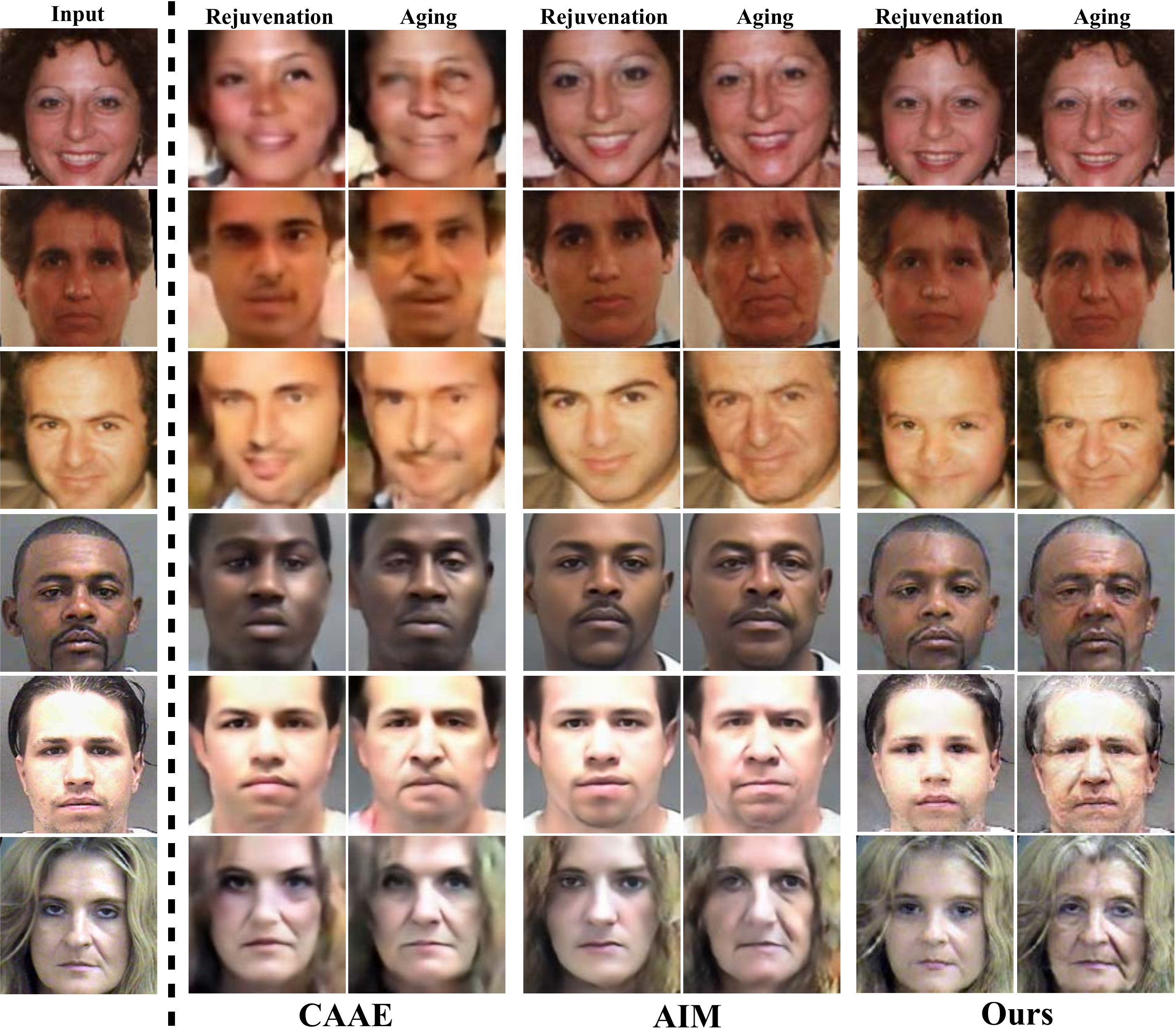}
    \caption{Qualitative comparisons with prior work on FG-NET~(top 3 rows) and MORPH~(bottom 3 rows).}
    \label{fig:qualitative_comparison}
\end{figure}

\subsection{Evaluation on GFR}

To validate the generalization ability of our \methodname for the general face recognition~(GFR), we further conduct experiments on the LFW~\cite{huang2008labeled} and MegaFace Challenge 1 Facescrub~(MF1-Facescrub)~\cite{kemelmacher2016megaface} datasets. LFW~\cite{huang2008labeled} is the most popular public benchmark dataset for GFR, which contains 13,233 face images from 5,749 subjects. The MF1-Facescrub~\cite{kemelmacher2016megaface} uses the Facescrub dataset~\cite{ng2014data} of 106,863 face images from 530 celebrities as a probe set. The most challenging problem of MF1 is that it uses an additional 1M face images in the gallery set to distract the face matching. That is, the results on MF1 are not as reliable as LFW due to the extremely noisy distractors in MF1. We strictly follow the same procedure as~\cite{wang2019decorrelated,wang2018orthogonal}, where the training dataset contains 0.5M images~(SCAF). Table~\ref{tab:gfr} reports the verification rate on LFW and the rank-1 identification rate in MF1-Facescrub against the state-of-the-art GFR methods. Our method achieves competitive performance on both datasets, demonstrating the strong generalization ability of our~\methodname. We highlight that our \methodname can provide photo-realistic synthesized faces to improve model interpretability, which is absent in other methods~\cite{wang2019decorrelated,wang2018orthogonal}.

\subsection{Evaluation on FAS}\label{sec:qualitative_comparison}

We further evaluate the model trained on SCAF for FAS. Fig.~\ref{fig:qualitative_results} presents some sample results on the external datasets including LCAF, MORPH, and FG-NET. Our method is able to simulate the face age synthesis process between age groups with high visual fidelity. Although there exist variations in terms of race, gender, expression, and occlusion, the synthesized faces are still photo-realistic, with natural details in the skin, muscles, and wrinkles while consistently preserving identities, confirming the generalization ability of the proposed method.

We conduct qualitative comparisons with prior work including CAAE~\cite{zhang2017age} and AIM~\cite{zhao2019look} on MORPH and FG-NET. Fig.~\ref{fig:qualitative_comparison} shows that both CAAE and AIM produce oversmoothed faces due to their image reconstruction while our \methodname uses the identity age condition to synthesize faces based on multi-level features extracted from the encoder. Note that the results of competitors are directly referred from their own papers for a fair comparison, which is widely adopted in the FAS literature such as~\cite{he2019s2gan,lihierarchical,li2019age,liu2019attribute,yang2018learning} to avoid any bias or error caused by self-implementation. 
See Appendix for quantitative comparisons with CAAE~\cite{zhang2017age}, IPCGAN~\cite{wang2018face}, and an ablation study of identity conditional module in terms of two evaluation criteria---age accuracy and identity preservation.
%!tex root=./cvpr.tex
\section{Conclusion}\label{sec:conc}

In this paper, we proposed a multi-task learning framework, termed \methodname, to achieve AIFR and FAS simultaneously. We proposed two novel modules: AFD to decompose the features into age- and identity-related features, and ICM to achieve identity-level face age synthesis.
Extensive experiments on both cross-age and general benchmark datasets for face recognition demonstrate the superiority of our proposed method.

\clearpage
{\small
\bibliographystyle{ieee_fullname}
\bibliography{ref}
}

%!tex root=cvpr.tex
\clearpage
\onecolumn
\appendix
\renewcommand \thepart{}
\renewcommand \partname{}
\part{\hfill \textsc{Appendix} \hfill}
\addcontentsline{toc}{section}{Appendix}
\numberwithin{equation}{section}
\setcounter{figure}{0}
\setcounter{table}{0}
 \renewcommand\thetable{\thesection.\arabic{table}}

\section{Quantitative comparison on face age synthesis} We trained all models on the SCAF dataset for fair comparisons, which are then directly applied to three external cross-age datasets; \ie, MORPH\cite{ricanek2006morph}, FG-NET~\cite{fgnet} and CACD~\cite{chen2015face}. We further evaluate them in terms of two widely-used metrics, \ie age accuracy and identity preservation, since the synthesized faces should be in the target age groups while consistently preserving the identities. 

These two metrics are detailed as follows:
\begin{enumerate}
    \item\textbf{Age accuracy:} We randomly select 80\% of LCAF to train a ResNet-100 model using $\ell_{\textsc{ae}}$ as the loss function to predict the ages of all synthesized faces and test the trained model on the remaining faces. Then, we compute the proportion of the predicted ages falling into the target age groups as the age accuracy.
    \item\textbf{Identity preservation:} An external well-trained face recognition model, ResNet-100 network pre-trained on MS-Celeb-1M dataset provided by~\cite{deng2019arcface}, is used for fair comparisons to compute the cosine similarity between the input and synthesized faces.
\end{enumerate}

Table~\ref{tab:results} presents the quantitative results of different face aging/rejuvenation methods, including CAAE~\cite{zhang2017age}, IPCGAN~\cite{wang2018face}, our proposed MTLFace and its variant~(w/o ICM). It can be observed that MTLFace outperforms CAAE and IPCGAN by a clear margin; this is a direct results of the AIFR task and ICM. On the other hand, without ICM, MTLFace reduces to a common cGANs-based method that uses one-hot encoding to control face aging/rejuvenation at the group level. Remarkably, the MTLFace without ICM still outperforms these two baseline methods, implying that our multi-learning framework with attention-based feature decomposition is effective in improving the age accuracy and identity preservation.

\begin{table*}[h]
    \centering
    \caption{Quantitative comparisons of our MTLFace with the state-of-the-art face aging/rejuvenation methods. We reported the mean value of age accuracy~(Age Acc.) and cosine similarity~(Cos. Sim.) computed over all age mappings. For each dataset, the best and second best  are  in \textcolor{red}{\textbf{red}}  and \textcolor{blue}{\textbf{blue}}, respectively.}
    \begin{tabular}{lcclcclcc}
    \toprule
    & \multicolumn{2}{c}{MORPH} && \multicolumn{2}{c}{FG-NET}& & \multicolumn{2}{c}{CACD} \\
    \cmidrule{2-3} \cmidrule{5-6} \cmidrule{8-9} 
    & Age Acc.~(\%) & \multicolumn{1}{l}{Cos. Sim.} & \multicolumn{1}{c}{} & Age Acc~(\%) & \multicolumn{1}{l}{Cos. Sim.} & \multicolumn{1}{c}{} & Age Acc.~(\%) & \multicolumn{1}{l}{Cos Sim.} \\ 
    \midrule
    CAAE~\cite{zhang2017age}    & 45.62      & 0.256      && 41.85   & 0.228   && 45.06  & 0.204    \\
    IPCGAN~\cite{wang2018face}  & 39.95      & 0.682      && 43.34   & 0.581   && 50.85  & 0.589    \\
    \hline
    
    MTLFace~(ours)              & \textcolor{red}{\textbf{57.40}}      & \textcolor{red}{\textbf{0.745}}      && \textcolor{red}{\textbf{61.47}}   & \textcolor{red}{\textbf{0.638}}   && \textcolor{red}{\textbf{60.62}}  & \textcolor{red}{\textbf{0.676}}    \\ 
    \quad\ w/o ICM          & \textcolor{blue}{\textbf{50.80}}     & \textcolor{blue}{\textbf{0.729}}      && \textcolor{blue}{\textbf{55.26}}   & \textcolor{blue}{\textbf{0.600}}   && \textcolor{blue}{\textbf{55.79}}  & \textcolor{blue}{\textbf{0.652}}    \\
    \bottomrule
    \end{tabular}
    \label{tab:results}
\end{table*}

\end{document}